%% file: main.tex
\documentclass{article}
\usepackage{spconf,amsmath,graphicx}
\usepackage[usenames,dvipsnames]{xcolor}
\usepackage[breaklinks=true,colorlinks,bookmarks=false, citecolor=Green]{hyperref}
\usepackage{url}
\usepackage{booktabs}
\usepackage{multirow}
\usepackage{changepage,threeparttable}
\usepackage{wrapfig}
\usepackage{pifont}
\usepackage{array}

\newcommand\offset{\mathit{offset}}

\title{Adapting CNNs for Fisheye Cameras without Retraining}
\name{Ryan Griffiths, Donald G. Dansereau \thanks{The authors are with the Australian Centre For Robotics (ACFR), School of Aerospace, Mechanical and Mechatronic Engineering, The University of Sydney, 2006 NSW, Australia.}}
\address{University of Sydney}
\begin{document}
\maketitle
\begin{abstract}
The majority of image processing approaches assume images are in or can be rectified to a perspective projection. However, in many applications it is beneficial to use non conventional cameras, such as fisheye cameras, that have a larger field of view (FOV). The issue arises that these large-FOV images can't be rectified to a perspective projection without significant cropping of the original image. To address this issue we propose Rectified Convolutions (RectConv); a new approach for adapting pre-trained convolutional networks to operate with new non-perspective images, without any retraining. Replacing the convolutional layers of the network with RectConv layers allows the network to see both rectified patches and the entire FOV.
We demonstrate RectConv adapting multiple pre-trained networks to perform segmentation and detection on fisheye imagery from two publicly available datasets. Our approach requires no additional data or training, and operates directly on the native image as captured from the camera. We believe this work is a step toward adapting the vast resources available for perspective images to operate across a broad range of camera geometries. Code and datasets are available at \href{https://roboticimaging.org/Projects/RectConv/}{https://roboticimaging.org/Projects/RectConv/}.
\end{abstract}
\input{sections/01_introduction}

\input{sections/02_related}
\input{sections/03_methods}
\input{sections/04_results}
\input{sections/05_conclusions}



\bibliographystyle{IEEEbib}
\bibliography{refs}

\end{document}

%% file: sections/01_introduction.tex
\section{Introduction}
\label{sec:intro}

Neural networks have found widespread adoption across a breadth of imaging tasks. 
However, adapting these networks to emerging imaging technologies generally requires gathering extensive new datasets reflective of new camera properties, even when the operating environment remains unchanged.
In this paper we propose a training-free approach that modifies pre-trained neural networks to operate with previously unseen cameras. 

We believe our approach could have applicability across a broad range of neural architectures and camera technologies. For this paper we focus on convolutional neural networks (CNNs) trained on conventional monocular imagery, and demonstrate adaptation to wide-field-of-view (FOV) fisheye-lens imagery. We show why adaptation is required and why our approach performs well where previous approaches like image rectification fail. We demonstrate our method with multiple tasks, networks, and cameras. 

CNN-based architectures generally assume translational invariance, in which features have similar appearance across the camera's FOV. However, many camera geometries including fisheye-lens cameras do not exhibit this invariance. CNNs trained on conventional imagery therefore perform poorly with fisheye-lens imagery, and in general applying CNNs across camera geometries yields degraded performance.%

\begin{figure}[t]
	\centering
	\includegraphics[width=0.8\linewidth]{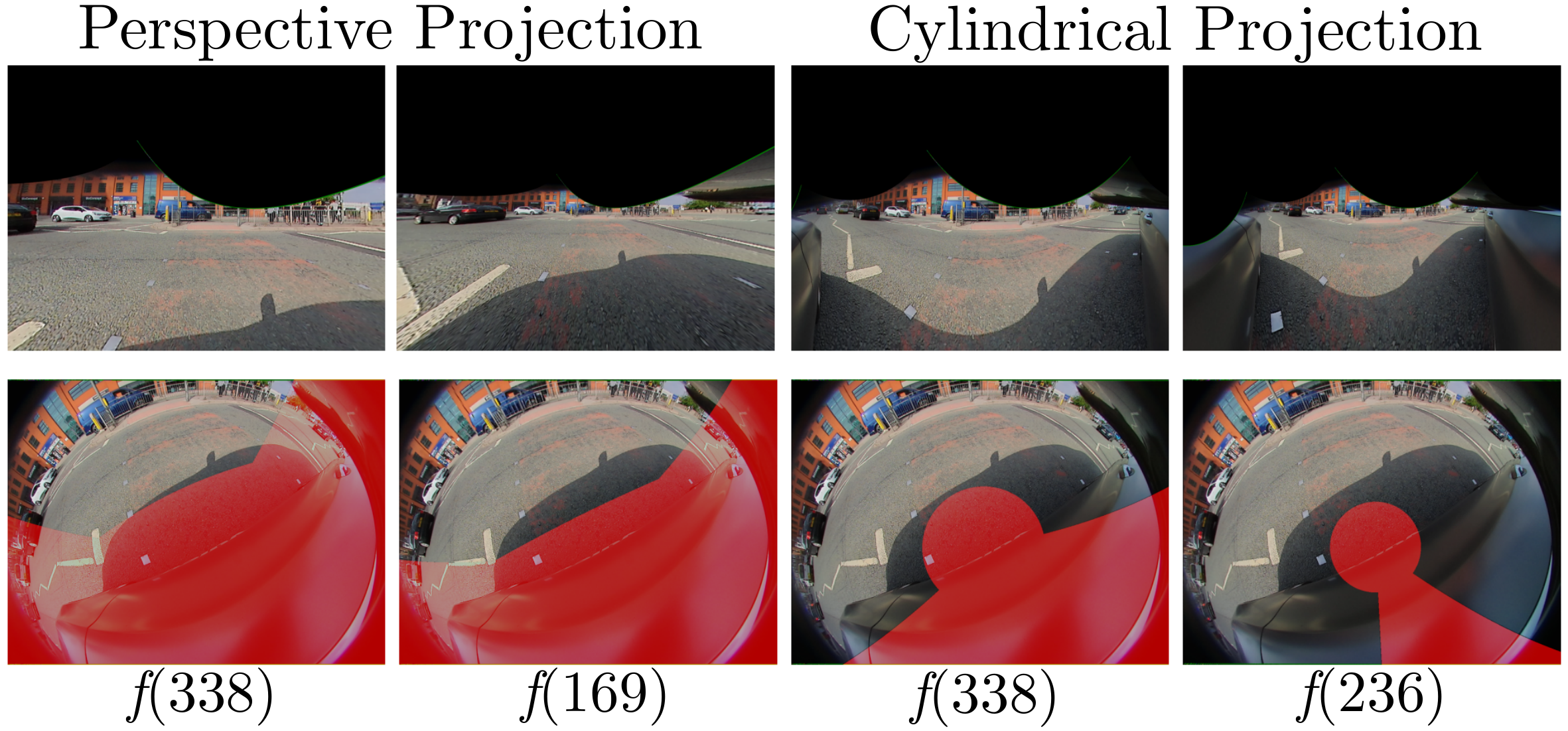}
	\caption{Example of perspective and cylindrical camera projections applied to a wide field of view fisheye image. Regions in red show areas that are excluded from the rectified projection. Decreasing the focal length can reduce cropping but increases distortion.}\label{fig:projections}
\end{figure}

Prior work has calibrated the camera and rectified its imagery such that translational invariance holds \cite{yogamani2019woodscape}. Alternatively, they rectify to a common equirectangular projection and adapt input convolutions to that geometry \cite{su2017learning}. 
However, no general mapping to a rectified image is possible without cropping and losing parts of the original image~\cite{courbon2007generic} (see Fig.~\ref{fig:projections}). Such methods therefore do not address all imaging geometries including fisheye. 

Patch-wise methods overcome some of these limitations but incur additional computational cost as multiple projections and inferences are required. Through tuning of patch parameters such methods trade computation against accuracy, incurring extensive computational cost for high-fidelity results~\cite{su2017learning}. They also only address local deformation.

\begin{figure}[t]
    \centering
    \includegraphics[width=0.7\linewidth]{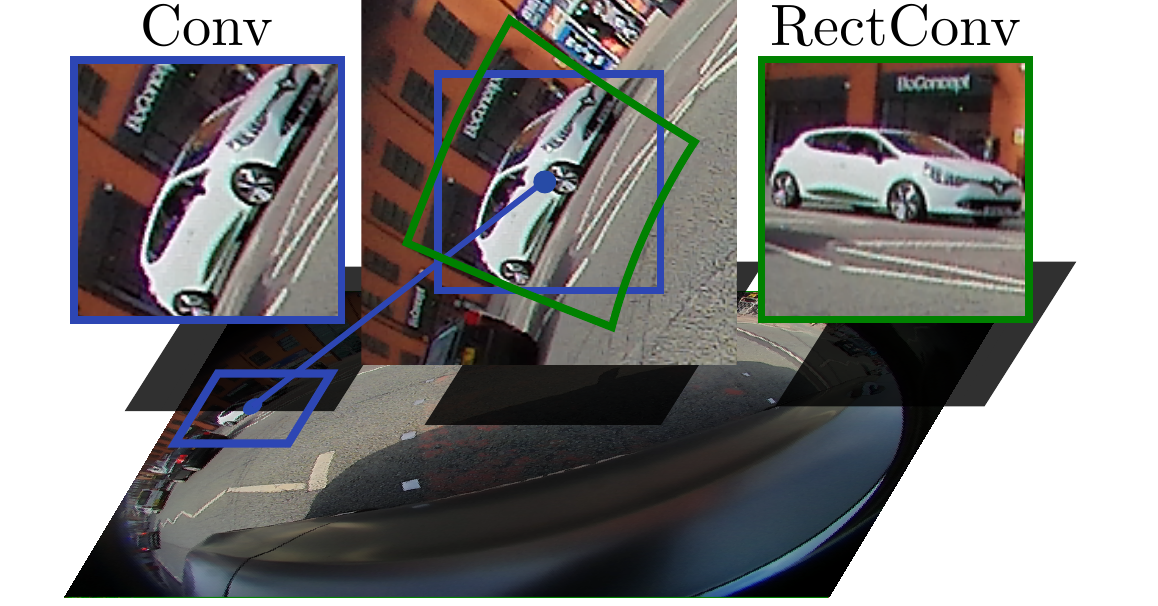}
    \caption{An illustration of what regular convolution and RectConv sees for a fisheye image at a given position in the image. Blue and green boxes indicate the kernel shapes for regular convolution and RectConv, respectively.}
    \label{fig:patch_rectification}
\end{figure}

Our approach introduces a modified convolutional layer called rectified convolution (RectConv) based on deformable convolutions \cite{dai2017deformable} and spherical convolutions \cite{su2017learning}. As depicted in Fig.~\ref{fig:patch_rectification}, rather than adapting  the image to the network, we adapt the kernel shape to the image geometry.

Replacing normal convolution with RectConv layers allows pre-trained networks to operate on new imaging geometries with improved performance.  To address both local and local deformation, we show RectConvs can be applied throughout the network, not only in the input layer. The only additional information required is a calibrated model of the camera, from which the RectConv deformations are computed.

The main contributions for this work are:
\textbf{(1)}~We propose RectConv convolution layers that allow networks to natively handle previously unseen camera geometries without requiring retraining or re-projection of input imagery;
\textbf{(2)}~We develop an approach for automatically adapting existing networks to RectConv networks, allowing pre-trained networks to be applied with new cameras; and
\textbf{(3)}~We compare with naive and rectification-based methods, showing improved performance for wide-FOV images on multiple networks architectures, cameras, and tasks.

Upon paper acceptance we will make all code available. We believe this work will allow efficient deployment of existing solutions with a breadth of existing and emerging camera technologies. While we focus on large-FOV cameras and fully convolutional neural networks, we anticipate extension to other network architectures and camera geometries is feasible.

%% file: sections/02_related.tex
\section{Related Work}
\label{sec:related}

\textbf{Large-FOV cameras.} Much research has gone into using large-FOV images, such as fisheye \cite{rashed2021generalized} and panorama \cite{ yang2019pass}. These larger fields of views can be hugely beneficial for specific applications like autonomous driving \cite{ye2020universal}. One common technique is to transform an existing perspective dataset to look like a large-FOV image e.g.~fisheye \cite{kim2022expandable, chen2023fisheye}, to aid in the training process. This allows existing datasets of conventional perspective images to be used, but requires retraining per camera geometry and does not entirely capture the target domain behaviour.

While there is a trend toward transformer-based architectures, CNN-based methods remain state of the art for many fisheye applications \cite{ye2020universal}.
These approaches require extensive datasets and training for the specific type of camera being used. This can be prohibitive, fails to benefit from the extensive existing pre-trained networks and datasets, and limits generalisation to different cameras.

\textbf{Adapting convolutions.} There are multiple works that aim to adapt convolutional layers to better suit a specific camera. Jaberberg et al.~\cite{jaderberg2015spatial} were among the first to adapt convolutions to learn spatial transformations, with their Spatial Transformer Networks (STNs). Follow-on work proposed Active Convolutions~\cite{jeon2017active} and applied a learnt offset, however this work only applied a single offset across the whole image, failing to address local deformation.

Deformable Convolutions~\cite{dai2017deformable, zhu2019deformable} are a more general version of active convolutions which learn an offset field mapping position in the image to local deformation. This makes it much more general at the cost of an increased number of learned parameters. Our work builds directly upon deformable convolutions where we employ camera calibration to derive a closed-form offset field to match the geometry of the input imagery, allowing us to operate directly on distorted images.

Facil et al.~\cite{facil2019cam} introduce camera-aware convolutions which embed the camera parameters into the feature maps of the CNN. 
This approach addresses conventional pinhole camera geometry and it is unclear whether it would generalise to other camera geometries. In contrast the proposed approach explicitly and efficiently addresses non-perspective camera geometries including fisheye.

\textbf{Spherical convolutions.} The line of work which is most closely related to ours is spherical convolution~\cite{su2017learning} and the many follow on works~\cite{coors2018spherenet, esteves2018learning}. This work applies CNNs trained on perspective images to 360\textdegree\ images. Zhang et al.~\cite{zhang2022bending} extend this to use transformers instead of CNNs with a focus on panoramic images, and Cohen et al.~\cite{cohen2018spherical} employ the Fast Fourier Transform (FFT) with increased speed and rotational invariance. However, in our case we don't want rotational invariance as rotation can be informative~\cite{su2019kernel}. These approaches require additional training or fine tuning, whereas our goal is to adapt to new cameras without retraining.

Tateno et al.~\cite{tateno2018distortion} who introduced Distortion-Aware Convolutions and Su et al.~\cite{su2019kernel} who introduced Kernel Transformer Networks, both have similar goals to ours in that they seek efficient adaptation of existing models from perspective imagery without retraining. They however focus on images given in an equirectangular projection specific to 360\textdegree\ imagery that does not generalise well to other camera geometries. As discussed by Yogamani et al.~\cite{yogamani2019woodscape} ``spherical models do not provide an accurate fit for fisheye lenses and it is an open problem''. Our approach doesn't require an equirectangular projection and can handle many imaging geometries including fisheye images.

%% file: sections/03_methods.tex
\section{Rectified Convolutions}
\label{sec:methods}
    

\subsection{RectConv Layers}
We propose an adaptation of the convolutional layer which we call RectConv. A RectConv layer does not have a fixed kernel shape, instead the shape matches the local deformation at the point in the image that the kernel is being applied to. This local deformation results in kernel ``offsets'' which are calculated based on how the patch would be rectified. Fig.~\ref{fig:patch_rectification} shows an example of the RectConv kernel shape and the corresponding view observed from that kernel. This adaptation of the convolutional layer is based on deformable convolutions, which provide a general framework for warping kernel shapes for each pixel location in an image. 
To achieve this, we require a way to calculate the local kernel offsets required for each pixel based on calibrated camera parameters. The offsets also need to be adjusted for each different network layer, especially for layers that modify size, such as pooling.

\textbf{Camera Model.}
Our method requires an invertible camera model that projects image points into 3D points of intersection with some reference surface in 3D space. 
A general form for such a model is $p  = f_{3D}(u, v)$, where $p$ is a point in space in Cartesian coordinates $(x,y,z)$ and $u, v$ are coordinates on the image plane. The function $f_{3D}$ is the camera projection from 2D to 3D. The inverse form is also required, $u,v = f_{2D}(p)$, where $f_{2D}$ converts a point in 3D space to its corresponding 2D image coordinates. Models of the required form are readily available for a broad range of cameras including the fisheye-lens cameras used in this work.

\begin{figure}[t]
	\centering
	\includegraphics[width=\linewidth]{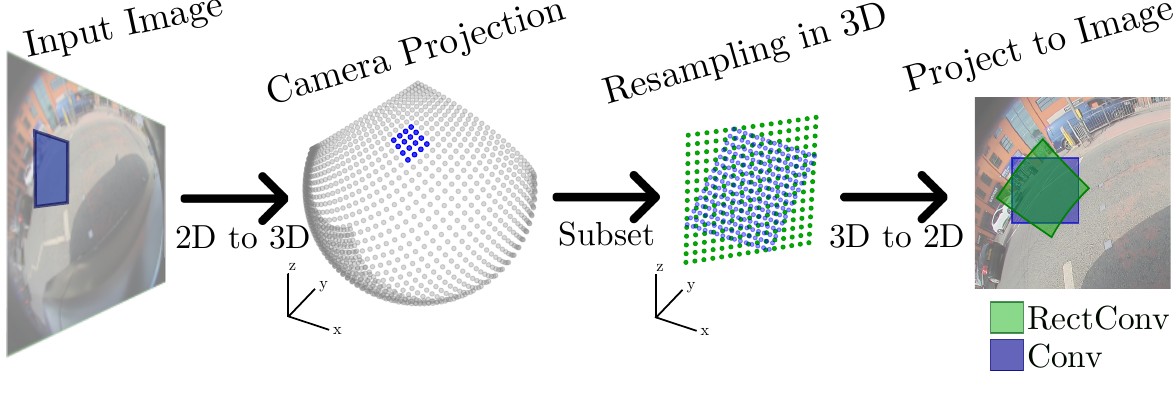}
	\caption{For a given patch each pixel is converted to 3D space which is then sampled on a regular planar grid. This grid in 3D space is converted back to image locations that represent the kernel locations for that position.}\label{fig:offsets}
    \vspace{-4mm}
\end{figure}

\textbf{Calculating Kernel Offsets.}
Here we derive the kernel offsets required at each image location. The process is depicted graphically in Fig.~\ref{fig:offsets}. The first step is to convert kernel pixels to points of intersection with a reference surface in 3D space,  
\begin{equation}
    p_i = f_{3D}(u_i, v_i),
\end{equation}
where $i$ denotes the different positions in the kernel. 
The scale of the patch in 3D space is computed as 
\begin{equation}
    s = \frac{w_{grid}+h_{grid}}{2},
\end{equation}
where $w_{grid}, h_{grid}$ are the horizontal and vertical size of the original grid and are calculated as $p_{max} - p_{min}$ in their respective dimensions. 
A linear planar sampling $k$ of scale $s$ is calculated to be tangential to the point $p_c$ at the centre of the original grid, from which new sample points can be calculated as
\begin{equation}
    \hat{p}_i = p_{c} + k_i.
\end{equation}
Here $\hat{p}_i$ is the new point in space at position $i$ in the kernel. With the new list of points in 3D space that can be converted back to the image plane, 
\begin{equation}
    \hat{u}_i, \hat{v}_i = f_{2D}(\hat{p}_i),
\end{equation}
where $\hat{u}_i, \hat{v}_i$ are the rectified pixel location on the image that the convolution should sample. For use within our framework these points need to be converted to the form 
\begin{equation}
    {\offset}_{i} = (\Delta u_i, \Delta v_i),
\end{equation}
where $\Delta u = \hat{u}_i - u_i$ and $\Delta v = \hat{v}_i - v_i$.
This $\offset$ value needs to be calculated for every position in the kernel and every position in the image. To reduce computation time we compute offsets for a subset of image locations and interpolate. As the cameras used had a continuous smooth projection, this effectively reduces computation without affecting performance.

\textbf{Modifying Offset Fields.}
The pre-calculated offset field needs to be modified depending on the parameters used in the convolutional layer. A layer with no padding results in a truncated output that can be accommodated by cropping the offset field to match the output size. A layer with stride can be accommodated by adjusting the sampling of the offset field to match the stride. Dilation results in an effective increase in kernel size, so an offset of a larger kernel should be used then sampled in the same way as the dilation process.

Finally, convolutional layers after a dimensional reduction, such as from a pooling layer or a previous convolution with stride, are accommodated by scaling the offset field in both size and magnitude. As the input size is scaled the offset values should be scaled by the same amount.

\textbf{Conversion from Conv to RectConv Layers.}
Conversion from a conventional CNN network to a RectConv version can be carried out efficiently and elegantly. Given a pre-trained model and camera parameters, a recursive search through the network modules identifies all the convolutional layers and replaces them with a RectConv layer. Offsets for the RectConv are computed as in the previous section, and weights and bias terms from the pretrained network are left unmodified. All convolution layers with a kernel size greater than one are converted to a RectConv layer in this manner. Layers with a kernel size of one require no modification and are left unchanged.

\subsection{Effects of Interpolation}
\label{sec:interp}
A consequence of employing non-integer offset fields in the deformable convolutions is that samples must be interpolated from the input imagery. Our implementation employs bilinear interpolation~\cite{dai2017deformable}. This process is information destroying and is present in every RectConv layer. The slight error at each layer accumulates as it propagates through the network. 

\begin{figure}[t]
    \centering
    \includegraphics[width=0.45\textwidth]{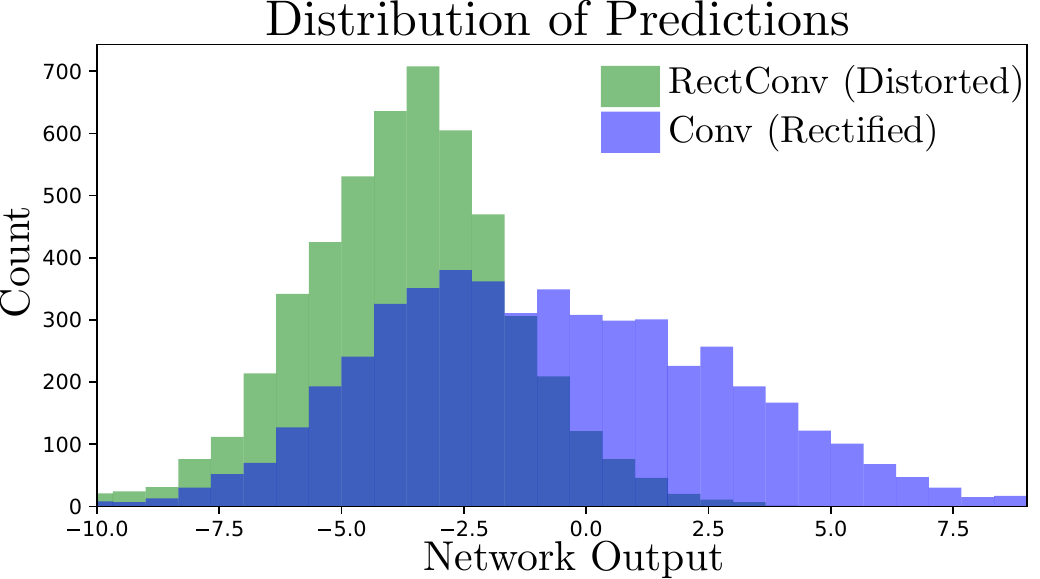}
    \vspace*{-2mm}
    \caption{A histogram of the outputs from a binary classification task showing how a RectConv layers result in a bias shift in the outputs.}\label{fig:bias}
    \vspace*{-4mm}
\end{figure}

Fig.~\ref{fig:bias} illustrates the effect of interpolation on the output of a network. This experiment shows a histogram of a binary classification network's output before a final classification layer is applied. The network used for this demonstration was a simple CNN, which has 4 convolutional layers with kernel sizes of 7, 5, 5 and 3, with no padding or stride. Then 4 additional $1\times1$ convolutional layers. The network was trained on a binary cats and dogs dataset.
The figure compares the convolutional form of the network applied to rectified perspective imagery, and the RectConv version applied to a distorted version of the same images. For an ideal conversion between convolutional and RectConv networks the outputs would be identical. However a shift in the distribution is evident, and we hypothesise arises due to the compounded impact of interpolation in the RectConv approach.

While these results show RectConv conversion is imperfect, it nevertheless demonstrates competitive performance in adapting to new camera geometries without a need for retraining. We leave further exploration and mitigation of the impact of interpolation as future work.

\subsection{Supported Model Architectures}
\label{sec:usable_models}

RectConv layers can be applied to any convolutional layer. However, there are some criteria needed for the model architecture to be a strong candidate for RectConv conversion. The model should not have a fixed input image size, instead it should be able to accept an image of arbitrary size. This is required as the perspective images use the train the model will not be the same size as the target (e.g. fisheye) images that will be used for inference. One architecture family with this behaviour is fully convolutional networks.


Given these considerations we chose to demonstrate our approach using fully convolutional networks ~\cite{long2015fully}. These architectures have seen success on a wide range of computer vision tasks. They do not have any fully connected layers and natively accept images of different sizes. We have also chosen networks that have no deconvolutional operations. While we do not demonstrate adaptation of deconvolutional layers to a rectified alternative, we believe generalisation is feasible and leave this as future work. 

\subsection{Fine-Tuning}
For this work we are explicitly interested in how networks can be adapted to new cameras without any additional training. However, we acknowledge that performing fine-tuning could provide a performance boost in many applications and would likely alleviate some of the bias shift due to interpolation discussed in Sec.~\ref{sec:interp}. 

%% file: sections/04_results.tex
\section{Experiments}
\label{sec:experiments}

\textbf{Tasks.}
We believe the proposed approach is general and applicable across many vision tasks. There are however certain tasks which are better suited to RectConv conversion than others. Size-conserving and pixel-wise labelling tasks such as segmentation and depth estimation are a strong fit, and for this reason we chose segmentation to demonstrate the effectiveness of the approach.

More challenging tasks have outputs with different dimensions to the input. An key example is object detection for which the outputs are a list of bounding boxes with pixel locations. We chose this task as a more challenging example for RectConv networks. An interesting side effect of conversion to RectConv layers is that because the kernels only see rectified patches the bounding box extents have been rectified locally around the object being bounded. This necessitates an additional step in which the rectified box extents are projected back to the original input image.

\textbf{Datasets.}
We demonstrated our approach on imagery from three different cameras drawn from two separate datasets. Firstly, Woodscape~\cite{yogamani2019woodscape} is a multi-task, multi-camera fisheye dataset. Woodscape has four fisheye cameras deployed on a vehicle, with data collected throughout a city environment. We demonstrated results using two cameras which capture the diversity of imaging geometries present in the dataset. 

The second dataset we use is PIROPO~\cite{del2021robust}. This dataset tracks people moving around a room from both a perspective and omnidirectional camera across multiple sequences. We used only the omnidirectional camera and demonstrated both segmentation and detection. The ground truth data provided includes a single labelled point for each person.

These datasets provide calibration parameters and are described with a radial distortion modelled using a 4th order polynomial.


\textbf{Models.}  We demonstrated our approach adapting four different pre-trained segmentation models constructed from two different backbones, ResNet50 and ResNet101~\cite{he2016deep}, and three separate architectures, FCN~\cite{long2015fully}, DeepLabV3 and DeepLabV3+~\cite{chen2017rethinking}.
We used an FCOS ResNet50~\cite{tian2019fcos} for object detection. These are representative of standard models for segmentation and detection, while also having readily available pre-trained weights.

\textbf{Pre-trained Networks.} 
Each test required a pre-trained convolutional network to be converted to the RectConv version. 
For the Woodscape dataset all pre-trained models were trained on the Cityscape dataset~\cite{cordts2016cityscapes}. Segmentation was evaluated using only the classes present in both Cityscape and Woodscape. In the case of DeeplabV3+ we use a publicly available network pre-trained on Cityscape. For the PIROPO dataset~\cite{everingham2010pascal} all pre-trained models were trained on Pascal VOC. We used the pre-trained models supplied by pytorch which are readily available. For this test only the person class was used and the other available classes were ignored.

\begin{table*}[t]\centering
    \caption{Comparison of segmentation mean intersection over union (MIOU) and pixel accuracy for pre-trained models applied to fisheye imagery from the Woodscape dataset. Bold denotes the best result.}
    \label{tab:segmentation}
    \scriptsize
    \begin{tabular}{lrrrrrrrrrr}\toprule
    \multirow{2}{*}{Camera} &\multirow{2}{*}{Method} &\multicolumn{2}{c}{FCN(Resnet50)} &\multicolumn{2}{c}{FCN(Resnet101)} &\multicolumn{2}{c}{DeeplabV3(Resnet50)} &\multicolumn{2}{c}{DeeplabV3+(Resnet101)} \\\cmidrule{3-10}
    & &Pixel Acc &MIOU &Pixel Acc &MIOU &Pixel Acc &MIOU &Pixel Acc &MIOU \\ \midrule
    \multirow{4}{*}{Camera 1} &Conv(Distorted) &83.16 &24.90 &82.60 &24.90 &73.64 &22.68 &81.03 &22.76 \\
    &Conv(Rectify) &82.05 &27.01 &85.27 &28.13 &72.64 &19.38 &86.45 &29.02 \\
    &Conv(Patches) &87.55 &29.98 &88.57 &29.8 &84.31 &28.04 &\textbf{91.54} &\textbf{32.47} \\
    &RectConv(Ours) &\textbf{87.68} &\textbf{31.05} &\textbf{89.12} &\textbf{31.68} &\textbf{84.61} &\textbf{29.28} &89.56 &30.84 \\ \midrule
    \multirow{4}{*}{Camera 2} &Conv(Distorted) &84.64 &24.09 &84.97 &24.53 &73.12 &20.74 &79.77 &21.31 \\
    &Conv(Rectify) &83.56 &24.88 &85.14 &25.62 &77.81 &21.28 &86.21 &26.37 \\
    &Conv(Patches) &89.09 &27.84 &89.44 &27.66 &\textbf{85.72} &25.83 &\textbf{91.48} &\textbf{28.54} \\
    &RectConv(ours) &\textbf{89.59} &\textbf{28.75} &\textbf{89.68} &\textbf{28.83} &84.78 &\textbf{26.23} &89.20 &27.17 \\
    \bottomrule
    \end{tabular}
    \vspace{-5mm}
\end{table*}

\subsection{Results}

\begin{figure}[t]
    \centering
    \includegraphics[width=\columnwidth]{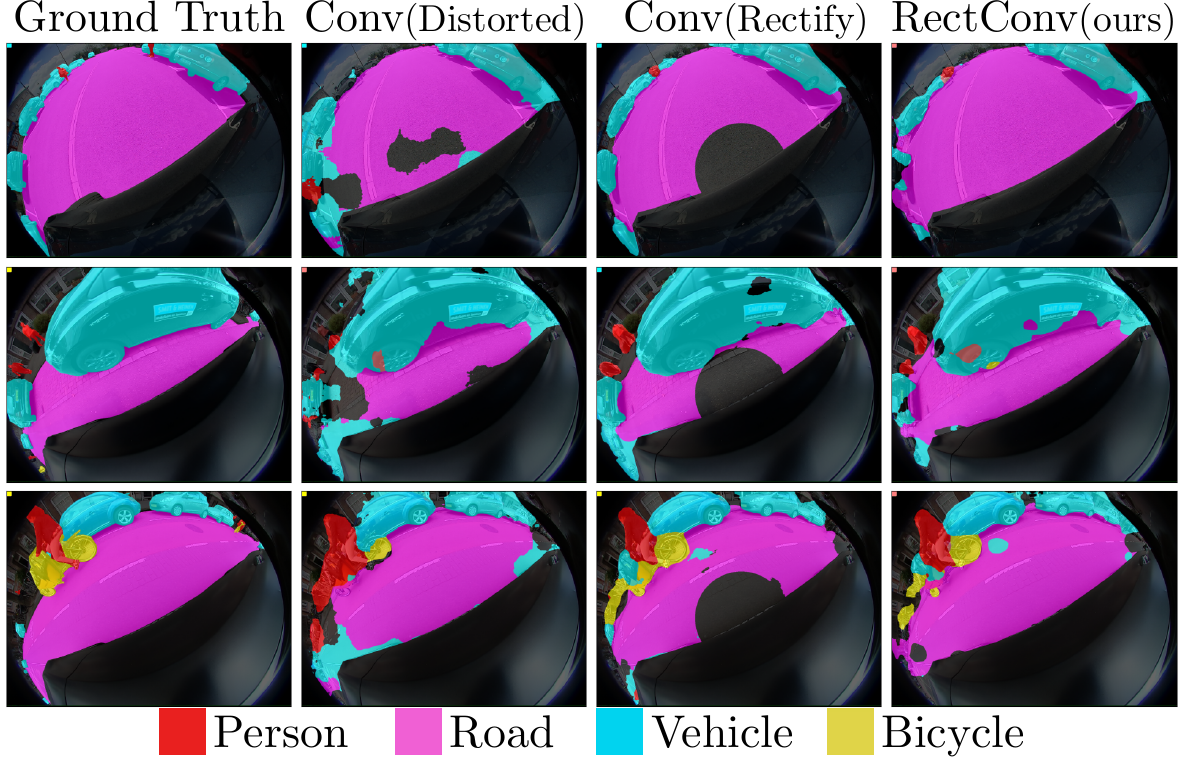}
    \caption{Comparison of segmentation using a FCN-Resnet101 pre-trained on Cityscape. The unmodified pre-trained network shows poor performance, pre-rectification shows poor performance and suffers from dead zones that could not be included in the rectification, and the proposed RectConv shows the strongest performance while covering the entire image.}
 \label{fig:segmentation}
 \vspace{-4mm}
\end{figure}

\textbf{Woodscape.} 
Tab.~\ref{tab:segmentation} and Fig.~\ref{fig:segmentation} show quantitative and qualitative results for the Woodscape dataset. 
We compared our method to three baseline approaches: the naive method of applying the pre-trained network directly to the distorted fisheye image; pre-rectifying the input images before inference using a cylindrical projection, as this projection maintains a larger field of view compared to other projection; and a patch-based approach which splits the image into multiple patches and rectifies them individually then runs inference on each patch. The last of these required manual adjustment to our specific application, such as patch size, location, overlap. 

Our method outperforms the alternatives with stronger quantitative results, as well as stronger qualitative results that cover the entire FOV when compared to the Distorted and Rectify methods. Importantly, our approach requires no additional training, only a one-time closed-form conversion of the convolutional layers to their RectConv alternatives. The performance of the patch-based method is only slightly lower, as expected since rectifying input patches solves many of the issues of the other approaches. The main drawback of the patch-based method is the significant increase in inference time as discussed in Sec.~\ref{sec:related} and demonstrated in the evaluation of inference time below. 

Fig.~\ref{fig:segmentation} clearly illustrates the differences between approaches and how they compare to the ground truth. For the pre-rectification method the results are distorted back to the original image geometry, making the cropped dead zone clearly visible in the centre of each segmentation mask. The additional errors from using a distorted input are also clearly visible for the convolution method, especially looking at the road prediction. 

\begin{figure}[t]
    \centering
    \includegraphics[width=\columnwidth]{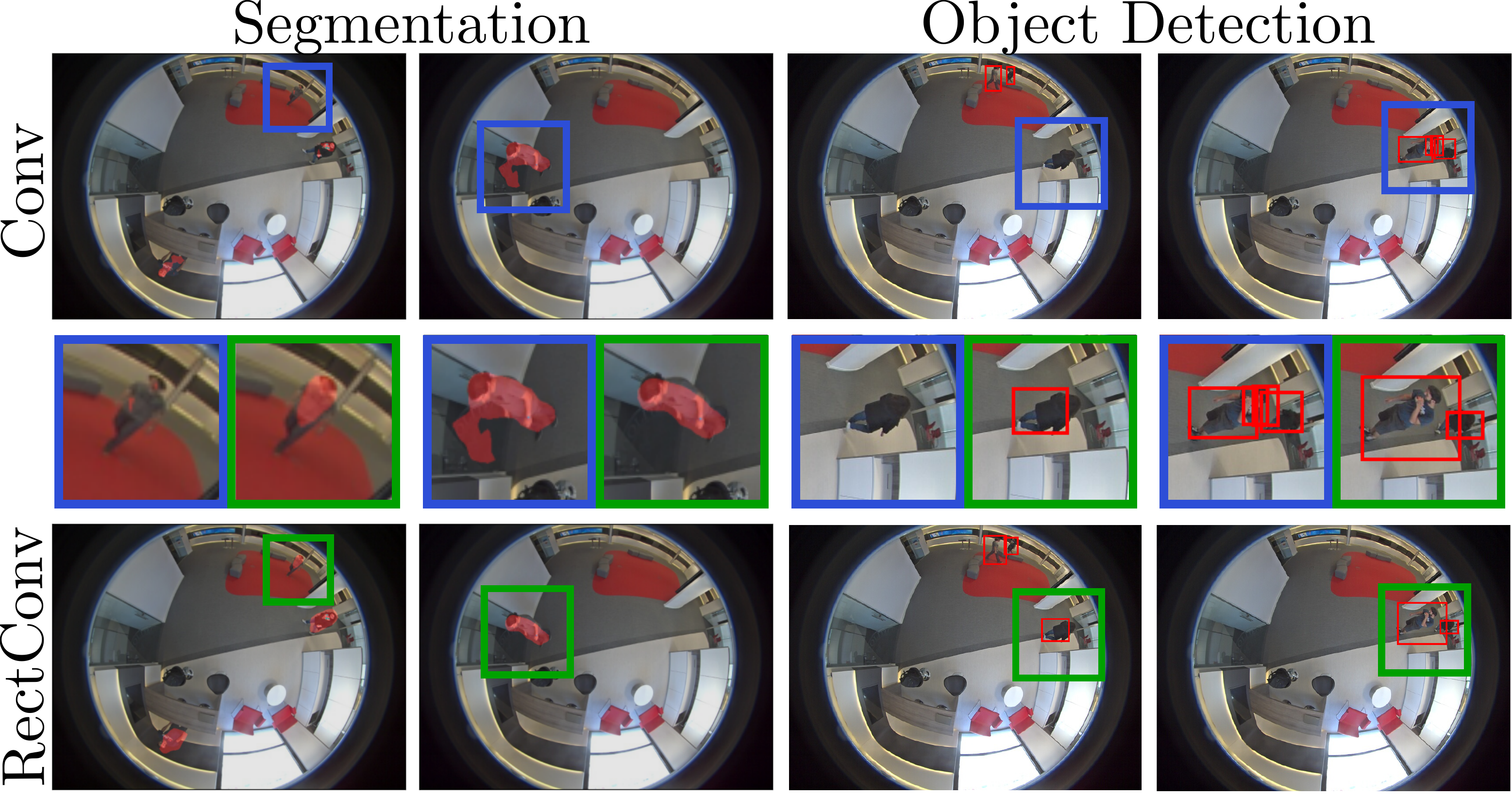}
    \caption{Detection and segmentation results for people on the PIROPO dataset \cite{del2021robust} using pre-trained segmentation (FCN-Resnet101) and object detection (FCOS-Resnet50) networks. RectConv has significantly improved the segmentation results and has a significant improvement in bounding box detections.}
 \label{fig:detection}
 \vspace{-5mm}
\end{figure}

\begin{table*}[t]\centering
    \caption{Comparison of segmentation and detection using pre-trained models on fisheye imagery from the PIROPO dataset.}
    \label{tab:detection}
    \scriptsize
    \begin{tabular}{lrrrrrrrrrr}\toprule
    \multirow{3}{*}{Method} &\multicolumn{6}{c}{Segmentation Network} &\multicolumn{3}{c}{Object Detection Network} \\\cmidrule{2-10}
    &\multicolumn{3}{c}{FCN(Resnet101)} &\multicolumn{3}{c}{DeeplabV3(Resnet101)} &\multicolumn{3}{c}{FCOS(ResNet50)} \\\cmidrule{2-10}
    &Precision &Recall &F1 Score &Precision &Recall &F1 Score &Precision &Recall &F1 Score \\\midrule
    Conv(Distorted) &69.09 &32.90 &44.57 &64.10 &28.86 &39.80 &84.64 &63.64 &72.65 \\
    RectConv(ours) &\textbf{76.91} &\textbf{48.05} &\textbf{59.15} &\textbf{81.00} &\textbf{44.30} &\textbf{57.28} &\textbf{86.67} &\textbf{65.66} &\textbf{74.71} \\
    \bottomrule
    \end{tabular}
    \vspace{-5mm}
\end{table*}

\textbf{PIROPO.}
Results for PIROPO are shown in Tab.~\ref{tab:detection} and Fig.~\ref{fig:detection}. As there is no ground truth segmentation available for this dataset the quantitative results are shown as an accuracy of correct detection of people within the image. These results show that not only does converting layers to RectConv increase true detection, it also reduces spurious detection. From the qualitative results we can see that the segmentation masks are cleaner compared to the naive approach.

In object detection, RectConv outperforms the conventional approach in all measures. The extent of the quantitative benefit is not as strong in detection as it is in segmentation. Conversion of bounding boxes in the RectConv network into image space is imperfect and is a topic for future research. Typical errors from the bounding box conversion are seen in Fig.~\ref{fig:detection}. 

\textbf{Inference Time}
The conversion of a network to use RectConv layers incurs additional inference-time computational cost due to the additional step of deforming the kernels. Tab.~\ref{tab:inference_time} shows the average inference time for the four networks we evaluated operating on a single image, running on an NVIDIA RTX3060, and the percentage increase in time compared to standard convolutions. Across the four different models there was an average of 180\% increase time for using the patch-based method. This was the best case scenario for the patch method, where only the relevant patches in the image were selected, significantly reducing the number of patches used.
There was a 60\% increase in time for using RectConv method, while slightly outperforming the patch-based method. We also expect that given optimisation the overhead required to run our method could be reduced.

\begin{table}[t]
\centering
\caption{Inference times using different methods. The \% increase compared to the standard convolutions is shown.}
\label{tab:inference_time}
\scriptsize
\begin{tabular}{lrrrr}\toprule
\multirow{2}{*}{Model} &\multicolumn{3}{c}{Inference Time Seconds - (\% Increase)} \\\cmidrule{2-4}
&Conv(Distorted) &Conv(Patches) &RectConv(Ours) \\\midrule
FCN Resnet50 &0.30 &0.83 (177\%) &0.46 (53\%) \\
FCN Resnet101 &0.41 &1.15 (180\%) &0.66 (60\%) \\
DeeplabV3 Resnet50 &0.32 &0.91 (182\%) &0.55 (71\%) \\
DeeplabV3+ Resnet101 &0.25 &0.69 (179\%) &0.38 (52\%) \\
\bottomrule
\end{tabular}
\vspace*{-5mm}
\end{table}

\subsection{Ablation Study}
\label{sec:ablation}

Tab.~\ref{tab:ablation} shows the results of an ablation study on how converting different parts of the network to RectConv layers affects overall performance. This study was performed on the Woodscape dataset, using the FCN ResNet101 segmentation network. It can be seen that the vast majority of the performance gain comes from converting the backbone. 
As the backbone is the part that is extracting geometric features it is understandable why it benefits most from RectConv. This supports the potential of the proposed approach to generalise well to other applications as these backbones are used for a range of tasks. 

\begin{table}[t]\centering
    \caption{Effect of different RectConv layers}\label{tab:ablation}
    \scriptsize
    \centering
    \begin{tabular}{ >{\centering\arraybackslash}m{0.7in}  >{\centering\arraybackslash}m{0.7in} >{\centering\arraybackslash}m{0.7in} >{\centering\arraybackslash}m{0.7in}}\toprule
    RectConv Backbone &RectConv Classification Head &Pixel Acc &MIOU \\\midrule
    \ding{55} &\ding{55} &82.60 &24.90 \\
    \ding{51} &\ding{55} &88.14 &30.01 \\
    \ding{55} &\ding{51} &82.63 &24.92 \\
    \ding{51} &\ding{51} &\textbf{89.12} &\textbf{31.68} \\
    \bottomrule
    \end{tabular}
    \vspace*{-2mm}
\end{table}



%% file: sections/05_conclusions.tex
\section{Conclusions}
\label{sec:concl}

We introduced RectConv, a training-free method for adapting pre-trained networks to new imagery. Our method operates directly on the new images without any pre-processing, additional data or fine-tuning. 
We demonstrated our approach by adapting segmentation and detection networks trained on conventional imagery to work with fisheye images. RectConv outperforms direct application of pre-trained networks and naive rectification.

In future we aim to demonstrate RectConv on additional tasks such as depth and pose estimation, with additional camera geometries. There is scope to expand the network conversion process to handle additional layer types like deconvolution, and as discussed in the Results section to improve performance around bounding box detections.